\documentclass[11pt]{article}
\usepackage{amsmath,amssymb,amsfonts,latexsym,epsfig,graphicx} 
\usepackage{algorithm,algorithmic}
\usepackage{psfrag,xspace} 
\usepackage{pgf} 
\usepackage{theorem} 

\newtheorem{theorem}{Theorem}

\newtheorem{definition}{Definition} 

\def\CC{{\mathbb C}} \def\NN{{\mathbb N}} \def\RR{{\mathbb R}} 
\usepackage{array}
\let\mathbfclassic\mathbf
\renewcommand{\mathbf}[1]{{\mathbfclassic{#1}}} 
\renewcommand{\algorithmiccomment}[1]{}
\algsetup{indent=0.5cm}

\begin{document}
\title{The assembly modes of rigid 11-bar linkages}

\author{ \begin{tabular}{ccc}
Ioannis Z.\ Emiris\thanks{emiris@di.uoa.gr}    &
	Guillaume Moroz\thanks{guillaume.moroz@inria.fr} & \\
National \& Kapodistrian University of Athens\qquad &
	INRIA Nancy & {\,}\hspace*{2cm}{\,} \\
Greece & France & \\
    \end{tabular}
}
\maketitle

\begin{abstract}
Designing an $m$-bar linkage with a maximal number of assembly modes
is important in robot kinematics, and has further applications in
structural biology and computational geometry.
A related question concerns the number of assembly modes of rigid mechanisms
as a function of their nodes $n$, which is uniquely defined given $m$.
Rigid 11-bar linkages, where $n=7$, are the simplest planar linkages
for which these questions were still open.
It will be proven that the maximal number of assembly modes of such
linkages is exactly $56$.

The rigidity of a linkage is captured by a polynomial system derived from
distance, or Cayley-Menger, matrices.
The upper bound on the number of assembly
modes is obtained as the mixed volume of a $5\times 5$ system.
An 11-bar linkage admitting $56$ configurations is constructed using
stochastic optimisation methods.
This yields a general lower bound of $\Omega(2.3^n)$ on the number
of assembly modes, slightly improving the current record
of $\Omega(2.289^n)$, while the best known upper bound is roughly $4^n$.
Our methods are straightforward and have been implemented in Maple.
They are described in general terms illustrating the fact that
they can be readily extended to other planar or spatial linkages. 

This version (2017) typesets correctly the last figure~\ref{F_26configs}
so as to include all~28 configurations modulo reflection.

\end{abstract} 

\section{Introduction}

Rigid mechanisms (or linkages) constitute an old but still very active area
of research in mechanism and linkage theory,
e.g.\ \cite{C02,FL94,WH07,WH07b,W77} as well as computational geometry and
structural bioinformatics,
e.g.\ \cite{BS04,EmiMou99,JRKT01,Hav98,TD99}. 

A given linkage may be represented by a graph with edge set $E$
of lengths $l_{ij}\in\RR_{+}$, for $(i,j) \in E$. An assembly mode, or
Euclidean embedding, in $\RR^d$ is a mapping of its vertices to
a set of points in $\RR^d$, such that $l_{ij}$ equals
the Euclidean distance between the images of the $i$-th and $j$-th vertices,
for $(i,j) \in E$.  Euclidean embeddings
impose no requirements on whether the edges cross each other or not.
A linkage is (generically) {\em rigid} in $\RR^d$ if and only if,
for generic edge lengths it can be embedded in $\RR^d$
in a finite number of ways, modulo rigid motions.
A graph is {\em minimally rigid} if and only if
it is no longer rigid once any edge is removed.

\begin{figure}
\psfrag{V[1]}[Br][tc][3][-137]{\raisebox{0.6em}{$$1\ \ $$}}
\psfrag{V[2]}[Bl][tc][3][-137]{$$2$$}
\psfrag{V[3]}[Br][tc][3][-137]{\raisebox{0.4em}{$$3$$}}
\psfrag{V[4]}[Br][tc][3][-137]{\raisebox{-0.4em}{$$4$$}}
\psfrag{V[5]}[Bl][tc][3][-137]{\raisebox{-0.4em}{$$5$$}}
\psfrag{V[6]}[Bl][tc][3][-137]{\raisebox{-0.4em}{$$6$$}}
\psfrag{V[7]}[Br][tc][3][-137]{$$7\qquad$$}
\begin{center}
\includegraphics[angle=137,width=0.7\linewidth]{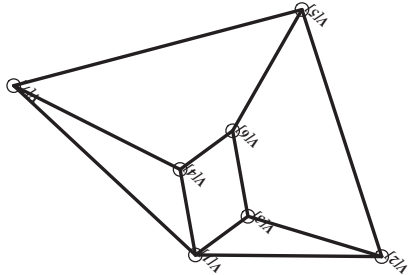}
\vspace{-6cm}
\end{center}
\caption{An 11-bar linkage. \label{11barLinkage}}
\end{figure}

Let us focus on planar linkages and the associated graphs.
A graph is called {\em Laman} if and only if the number of edges is $2n-3$,
where $n$ is the number of vertices and, additionally,
all of its vertex-induced subgraphs on $3 \le k<n$ vertices have $\le 2k-3$ edges.
This is essentially the Gr\"ubler-Kutzbach-Chebychev formula on
the degrees of freedom for mechanical linkages, e.g.~\cite{A89}.
It is a fundamental theorem that the class of Laman graphs coincides with the
generically minimally rigid planar linkages \cite{L70}. 
The problem studied in this paper is to compute the number
of distinct planar assembly modes of rigid mechanisms, up to rigid motions.
In particular, the maximal number of assembly modes of an $11$-bar linkage
(Figure~\ref{11barLinkage}) will be presented for the first time.

\subsection{Existing work}

The algebraic approach is to define a well-constrained polynomial system
expressing the length constraints, such that the system's real solutions
correspond precisely to the different assembly modes.  
When defining a straightforward system 
such as~(\ref{Esystem2d}), where the unknowns are the nodes' coordinates,
all nontrivial equations are quadratic.
For planar linkages,
there are $2n-4$ equations hence, by applying the classical B\'ezout bound
on the number of common roots, we obtain $4^{n-2}$.
It is indicative of the hardness of the problem that efforts to substantially
improve these bounds have failed.

Today, the best general upper bound is roughly
$ 
4^{n-2}/ \sqrt{\pi (n-2)} . 
$
This was obtained using 
determinantal varieties defined by distance matrices~\cite{BS04}.
Straightforward application of mixed volumes (discussed in
Section~\ref{Salgebra}) yields an upper bound of $4^{n-2}$ \cite{ST10}.

\begin{figure}[t]
\psfrag{V[1]}[Br][tc]{$$1$$}
\psfrag{V[2]}[Bl][tc]{$$\qquad2$$}
\psfrag{V[3]}[Br][tc]{$$3$$}
\psfrag{V[4]}[Br][tc]{$$4$$}
\psfrag{V[5]}[Bl][tc]{$$\qquad5$$}
\psfrag{V[6]}[Br][tc]{$$6$$}
\begin{center}
    \includegraphics[width=0.7\linewidth]{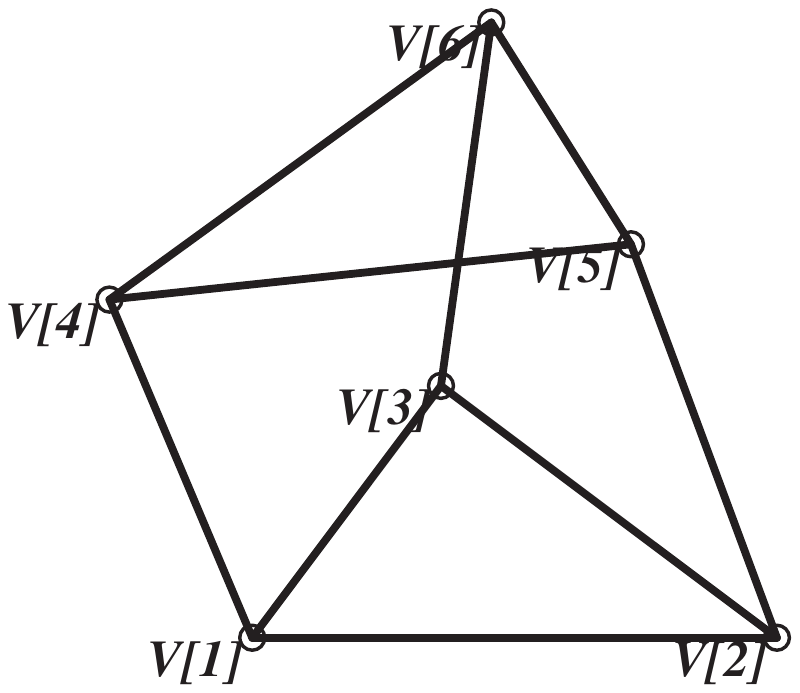}
\end{center}
\caption{The Desargues linkage or, equivalently, a planar parallel robot.}
\label{desarguesGraph}    
\end{figure}

The best general lower bounds are
$24^{\lfloor (n-2)/4 \rfloor}\simeq 2.21^n$ and
$2\cdot 12^{\lfloor (n-3)/3 \rfloor}\simeq 2.289^n/6$, obtained
by a caterpillar and a 
fan~\footnote{This slightly corrects the exponent of the original statement.}
construction, respectively~\cite{BS04}. 
Both bounds are based on the Desargues
graph (Figure~\ref{desarguesGraph}), which admits~24 assembly modes.

In applications, it is crucial to know the number of assembly modes
for specific (small) values of $n$.
The most important result in this direction was to show that the Desargues 
linkage, also known as the planar parallel robot,
admits precisely~24 assembly modes in the plane
\cite{BS04,H83,GSR92,LazMer94}.
Moreover, the $K_{3,3}$, or 9-bar, linkage admits exactly~16 assembly modes
in the plane~\cite{WH07}. This paper demonstrated such a linkage through
an elaborate construction which was not required in our case.

All existing bounds for $n\le 10$ are found in
Table~\ref{tab:2D-constructions} from \cite{EmTsVa09}
\footnote{see~\protect\cite{EmTsVa13} for an update.}.

\begin{table*}[ht]
  \centering  $ \begin{array}{c||c|c|c|c||c|c|c|c}
    n = & 3& 4 & 5 & 6 & 7 & 8& 9 & 10\\
    \hline \hline
    \mbox{upper}& 2& 4 & 8 & 24 & 64 & 128& 512 & 2048\\
    \hline \hline
    \mbox{lower} & 2& 4 & 8 & 24 & 48 & 96& 288 & 576 \\
    \hline \hline
\end{array} $ 
\medskip
\caption{Bounds for the number of embeddings of rigid graphs with $n\le 10$
\protect\cite{EmTsVa09}.} \label{tab:2D-constructions}
\end{table*}

\subsection{Our contribution}

To upper bound the number of planar rigid mechanisms, we explore
adequate polynomial systems leading to tight root bounds.
Modeling physical systems by appropriate polynomial systems 
is a deep and hard question, with a wide
range of applications in different fields.

We employ two powerful algebraic tools for defining polynomial
systems and for counting the system's common roots.
For the former, we use distance matrices, also known as Cayley-Menger
matrices, which contain all known and unknown distances between the
graph's nodes.
The signs of the matrix minors capture rigidity and embeddability
in Euclidean spaces, as described in Section~\ref{Salgebra}.
Our results indicate that such matrices are advantageous to using the
coordinates' formulation, such as system~(\ref{Esystem2d}),
when constructing the polynomial system.

Our second tool is the mixed volume of a well-constrained polynomial system,
which exploits the sparseness of the equations.
The mixed volume bounds the number of common roots, by
Bernstein's Theorem~\ref{Tbkk}, as described in Section~\ref{Salgebra}.
This bound is never larger than B\'ezout's, and is typically much
tighter for systems whose equations do not contain all possible terms
for a given total degree.
It turns out that such are the systems encountered here and
could be of wider interest in similar enumeration problems.  

Moreover, mixed volume is a bound so it only needs to consider which
are the nonzero coefficients, without considering specific values.
This is in sharp contrast to solving methods often used for root counting,
where one assigns random values to the coefficients.

We thus obtain a tight bound for linkages with $n = 7$ nodes by
appropriately formulating the polynomial system based on distance matrices.
We compute the mixed volume of a $5\times 5$ polynomial system, which equals 56.
This is tight since we demonstrate a construction with as many assembly modes,
see Figure~\ref{F_26configs}.
We also show how our stochastic optimisation methods are generalizable
for studying further planar or spatial linkages.
Our construction also yields a general lower bound of $\Omega(2.3^n)$ on
the number of configurations, thus improving the available bound of
$\Omega(2.289^n)$, whereas the best available upper bound is $O(4^n/ \sqrt{n})$.

The rest of the paper is structured as follows.
Section~\ref{Srigid} presents rigid mechanisms and how our study focuses
on a specific 11-bar linkage.
Section~\ref{Salgebra} presents our algebraic tools and
obtains the upper bound of 56 for $7$ nodes and 11 bars.
Section~\ref{Slower} actually constructs a linkage with 56 assembly modes.
We conclude with open questions.

\section{Rigid linkages}\label{Srigid}

This section studies the number of assembly modes of planar rigid linkages
or, equivalently, rigid graphs and their number of embeddings in $\RR^2$.

Such linkages admit inductive constructions that begin with a triangle,
followed by a sequence of so-called Henneberg steps.
There are two types of such steps, each adding one new vertex and
increasing the total number of edges by 2.
A graph is Laman, or the associated mechanism is rigid,
if and only if it can be constructed by a sequence of the corresponding
Henneberg steps.  
By exploiting this fact, all rigid graphs in $\RR^2$, but also in $\RR^3$,
were constructed using the Henneberg steps in~\cite{EmTsVa09}, and
classified up to graph isomorhism.
 
\begin{figure}[t]
\centering\includegraphics[width=0.7\columnwidth]{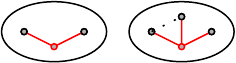}
\caption{The planar Henneberg steps; the bottom vertex is new.}
\label{planarHenneberg}
\end{figure}

Let us consider the Henneberg steps defining Laman graphs, each
adding a new vertex and a total number of two edges.
A Henneberg-1 (or $H_1$) step connects the new vertex to two existing vertices.
A Hennenerg-2 (or $H_2$) step connects the new vertex to
3 existing vertices having at least one edge among them,
and this edge is removed.
Both steps are illustrated in Figure~\ref{planarHenneberg}.
We represent each Laman graph by $\bigtriangleup s_4 \ldots,s_n$, where
$s_i \in\{1,2\}$; this is known as
its {Henneberg sequence}. Note that this sequence is by no means unique. 
A Laman graph is called $H_1$ if and only if it can be constructed
using only $H_1$ steps; it is called $H_2$ otherwise.  

Since two circles intersect generically in two points, a $H_1$ step at most doubles
the number of assembly modes and this is tight, generically.
In Table~\ref{tab:2D-constructions} most lower bounds follow from this fact.
The lower bound for $n= 9$ follows from the Desargues fan~\cite{BS04}.

For a Laman graph on~6 vertices, a tight uppper bound of~24 follows by examining
the only three possibilities: the graph is either $H_1$,
it is $K_{3,3}$, or is the Desargues graph. Now, $H_1$ graphs on~6 vertices
have exactly~16 assembly modes.
The $K_{3,3}$ graph has precisely~16 assembly modes~\cite{WH07},
a fact first conjectured in~\cite{W77}.
The Desargues graph has precisely~24 assembly modes:
the upper bound was first shown in~\cite{H83} and proven
more explicitly, along with the lower bound, in~\cite{GSR92}.
This fact was independently rediscovered in~\cite{BS04} and proven via coupler curves.

In~\cite{EmTsVa09} further upper bounds were obtained from the following
family of simple systems, employed in order to express graph
embeddability in $\RR^2$.
Let $x_i,y_i$ denote the coordinates of the $i$-th vertex,
where the $l_{ij}$ are the given lengths:
\begin{equation}\label{Esystem2d}  \left\{ \begin{array}{ll}
      x_i=a_i, \; y_i=b_i, \; & i=1,2, \quad a_i,b_i\in\RR, \\
      (x_i-x_j)^2+(y_i-y_j)^2= l_{ij}^2,\;\; & (i,j)\in E-\{(1,2)\}
    \end{array} \right.
\end{equation}
Vertex $(x_1,y_1)$ was fixed to $(0,0)$, to discard translations, and
$(x_2,y_2)$ to $(1,0)$ to remove rotations and scaling, assuming 
without loss of generality that the corresponding edge exists.
Then the mixed volume of the system gave the upper bounds in the table
which, unfortunately, are loose, even after removing roots at infinity
\cite{EmTsVa09}. Below we overcome this limitation by considering
polynomial systems derived from distance matrices.

We now study the case $n=7$.
If a $H_1$ step is applied to any linkage with $n=6$, the resulting
linkage with $n=7$ admits exactly 48 assembly modes, which is reflected
to the corresponding lower bound in Table~\ref{tab:2D-constructions}.
To maximize the number of assembly modes, we shall apply a
$H_2$ step to any graph with $n=6$. By checking graph isomorphisms and
taking into account the various symmetries, it was shown \cite{EmTsVa09}
that there are only~3 relevant graphs to be considered.
These are obtained by a $H_2$ step applied to the Desargues graph
as follows, where notation refers to Figure~\ref{desarguesGraph}:
\\
We remove edge $(4,5)$, and add edges $(4,7),(5,7)$, and
\begin{itemize}
\item $(1,7)$, or
\item $(3,7)$, or
\item $(6,7)$.
\end{itemize}

The first case corresponds to the topology that shall be studied extensively
in the sequel, since it leads to the maximum number of 56 assembly modes.
The linkage is shown in Figure~ref{11barLinkage} and
in the next section we will proove that 56 is
the maximum number of assembly modes possible for such a linkage.  
The other two linkages admit at most~44 and~48 assembly modes, respectively,
hence they are not studied any further.
These upper bounds are obtained as mixed volumes of polynomial systems,
as explained in Section~\ref{Salgebra}.  

\section{Algebraic methods for the upper bound} \label{Salgebra}

This section discusses polynomial systems, describes distance matrices,
and derives an upper bound on the number of assembly modes.

\subsection{Mixed volume}

We first discuss multivariate polynomial systems and introduce sparse elimination theory
in order to exploit their structure and sparseness; for details, see~\cite{CLO2,EmiPhd}.
Classical elimination theory characterizes every polynomial by its total degree.
For a well-constrained system of polynomial equations,
the classical B\'ezout bound on the number of isolated roots equals
the product of the polynomials' total degrees.
One disadvantage of this bound is that it counts projective roots and hence increases
when there are roots at projective infinity.

In sparse (or toric) elimination theory, a polynomial is characterized by
its support. 
Given a polynomial $f$ in $n$ variables, its support is the set of exponents
in $\NN^n$ corresponding to nonzero terms (or monomials).
The Newton polytope of $f$ is the convex hull of its support and lies in $\RR^n$.
Consider (Newton) polytopes $P_i\subset\RR^n$ and parameters
$\lambda_i\in\RR, \lambda_i\ge 0$, for $i=1,\dots,n$.
We denote by $\lambda_i P_i$ the corresponding scalar multiple of $P_i$.
%
Consider the Minkowski sum of the scaled polytopes
$\lambda_1 P_1+\cdots+\lambda_nP_n \in\RR^n$; its Euclidean volume
is a homogeneous polynomial of degree $n$ in the $\lambda_i$.
The coefficient of $\lambda_1\cdots \lambda_n$ is the
{\em mixed volume} of $P_1,\ldots,P_n$.
If $P_1 = \cdots =P_n$, then the mixed volume is $n!$ times the 
Euclidean volume of $P_1$.

We now focus on the topological {torus} $\CC^*=\CC-\{0\}$ in order to state
Bernstein's root bound in terms of mixed volume.
 
\begin{theorem} {\rm \cite{Bern75}}\label{Tbkk}
Let $f_1=\cdots= f_n=0$ be a polynomial system in $n$ variables
with real coefficients, where the $f_i$ have fixed supports.

  The number of isolated common solutions in $(\CC^*)^n$ is bounded above 
  by the mixed volume of the Newton polytopes of the $f_i$.
  This bound is tight for a generic choice of coefficients of the $f_i$'s.
\end{theorem}

One alternative to using general bounds such as mixed volume is to manipulate
a system, with coefficients chosen randomly, so as to bound the number of
common real roots, e.g.\ by means of Gr\"obner bases or homotopy continuation.
We actually used Gr\"obner bases for all systems studied in this paper,
where we have used random coefficients.
In particular, for the linkage analyzed below, we computed the
total-degree Gr\"obner basis of system~(\ref{Esystem2d}) with random distances
in Maple, and obtained the Hilbert polynomial of
the corresponding ideal.

This is a univariate polynomial where, by setting
its variable to 1, one deduces an upper bound on the number of complex
common roots.  This was indeed 56, which coincides with mixed volume.
However, this only offers an indication on the number of roots, not an actual
bound, since it depends on the choice of coefficients.

The advantage of mixed volume
is that it treats entire classes of systems defined by their nonzero terms,
without considering specific coefficient values.
Lastly, a tight mixed volume implies that one can solve the system
efficiently either by sparse resultants, e.g.\ \cite{CanEmi00},
or by sparse homotopies, e.g.\ \cite{Vers99}.

\subsection{Distance geometry}

The theory of distance geometry has been well developed, e.g.\
\cite{Blum70,Scho35}, 
with several applications especially
in structural bioinformatics, e.g.~\cite{EmiMou99,Hav98}. 

\begin{definition}\label{Dmatrix}
Given an $n$-vertex graph, the corresponding {\em distance} or
{\em Cayley-Menger matrix} is a symmetric $(n+1)\times (n+1)$ matrix $B$,
indexed by $i=0,\dots,n$,
such that $B(0,i)=1$ for $i\ge 1$, 
$B(i,i)=0$ for $i\ge 0$, and
$B(i,j)$ equals the squared distance between vertices $i,j$, for $j>i\ge 1$.
\end{definition}

To be more precise, to obtain the 
Cayley-Menger matrix of the given $n$ vertices, we must multiply $B$ by $-1/2$.
But for testing embeddability, the two formulations are equivalent, as seen in
the following theorem, due to the work of Cayley and Menger, see
\cite{Blum70,Scho35}.
Let $D(i_1,\dots,i_k)$ denote the $(k+1)\times (k+1)$ diagonal minor of $B$
indexed by rows and columns $0,i_1,\dots,i_k$, where
$i_1< \cdots <i_k\in \{1,\dots,n\}$.

\begin{theorem} \label{Trank}
Suppose we are given 
a matrix $B$ of the form specified in Definition~\ref{Dmatrix}.
Then $B$ corresponds to a (complete) graph embeddable in $\RR^d$,
if and only if
\begin{enumerate}
\item
for $k=2,\dots,d+1$ and any $\{i_1,\dots,i_k\}\subset\{1,\dots,n\}$,
it holds $(-1)^{k} D (i_1,\dots,i_k) \geq 0$, and
\item
rank$(B)=d+2$.
\end{enumerate}
\end{theorem}

The second condition, due to Cayley, yields a (large) number of equalities,
which are typically not independent.

The first condition, due to Menger, yields inequalities: 
For $k=2$, it expresses the fact that all entries must be non-negative.
For $k=3$, it captures the triangular inequality.
If we apply it, without loss of generality, to indices $1,2,3$, using
$c_{ij}=l_{ij}^2$ for the given entries, then the condition states that
$D(1,2,3)\le 0$, where this quantity equals
$$
- (l_{12}+l_{13}+l_{23})(l_{12}+l_{13}-l_{23})
(l_{12}+l_{23}-l_{13})(l_{13}+l_{23}-l_{12}) .
$$
This vanishes precisely when the corresponding points
are collinear, whereas $D(1,2,3) < 0$ when the points define a triangle.
Equivalently, $D(i,j,k)\le 0$ can be written 
$l_{ik} + l_{jk} \ge l_{ij}$ for all triplets $i,j,k\in\{1,\dots,n\}$.
For $k=4$ the condition captures the tetrangular inequality.

\subsection{Upper bound} 

We shall employ the rank condition 
in order to derive equality constraints on the unspecified distances of our linkage.
Here is the matrix, where the $c_{ij}=l_{ij}^2$ correspond to the
fixed distances, and $x_{ij}$ are the unspecified distances.
$$
\begin{array}{rl}
& \hspace*{4.5mm} \begin{array}{ccccccccc}
&\, v_{1} &\;\, v_{2} &\;\, v_{3} & \;\; v_{4} & \;\, v_{5} & \;\; v_{6} & \;\, v_7\end{array}
\\
\begin{array}{c}
\\ v_{1} \\ v_{2} \\ v_{3} \\ v_{4} \\ v_{5} \\ v_{6} \\ v_7\\
\end{array} & \left[ \begin{array}{cccccccc}
0 & 1 & 1        & 1            & 1       & 1            & 1 & 1    \\
1 & 0 & c_{12} & c_{13} & c_{14} & x_{15} & x_{16} & c_{17} \\
1 & c_{12} & 0 & c_{23} & x_{24} & c_{25} & x_{26} & x_{27} \\
1 & c_{13} & c_{23} & 0 & x_{34} & x_{35} & c_{36} & x_{37} \\
1 & c_{14} & x_{24} & x_{34} & 0 & x_{45} & c_{46} & c_{47} \\
1 & x_{15} & c_{25} & x_{35} & x_{45} & 0 & c_{56} & c_{57} \\
1 & x_{16} & x_{26} & c_{36} & c_{46} & c_{56} & 0 & x_{67} \\
1 & c_{17} & x_{27} & x_{37} & c_{47} & c_{57} & x_{67} & 0 \\
\end{array} \right]  \end{array}
$$
By Theorem~\ref{Trank}, any $5\times 5$ minor of this matrix must vanish,
which yields polynomial equations on the variables $x_{ij}$.
There are ${7\choose 3}=35$ such minors, each in~2 to~4 variables.
Among these polynomials,
no $4\times 4$ subsystem exists that corresponds to a rigid mechanism,
as can be verified by checking Laman's condition.
This is indispensable for the system to have a finite number of solutions.

\begin{figure}
\psfrag{V[1]}[Br][tc][3][-137]{\raisebox{0.6em}{$$1\ \ $$}}
\psfrag{V[2]}[Bl][tc][3][-137]{$$2$$}
\psfrag{V[3]}[Br][tc][3][-137]{\raisebox{0.4em}{$$3$$}}
\psfrag{V[4]}[Br][tc][3][-137]{\raisebox{-0.4em}{$$4$$}}
\psfrag{V[5]}[Bl][tc][3][-137]{\raisebox{-0.4em}{$$5$$}}
\psfrag{V[6]}[Bl][tc][3][-137]{\raisebox{-0.4em}{$$6$$}}
\psfrag{V[7]}[Br][tc][3][-137]{$$7\qquad$$}
\begin{center}
    \includegraphics[angle=137,width=0.7\linewidth]{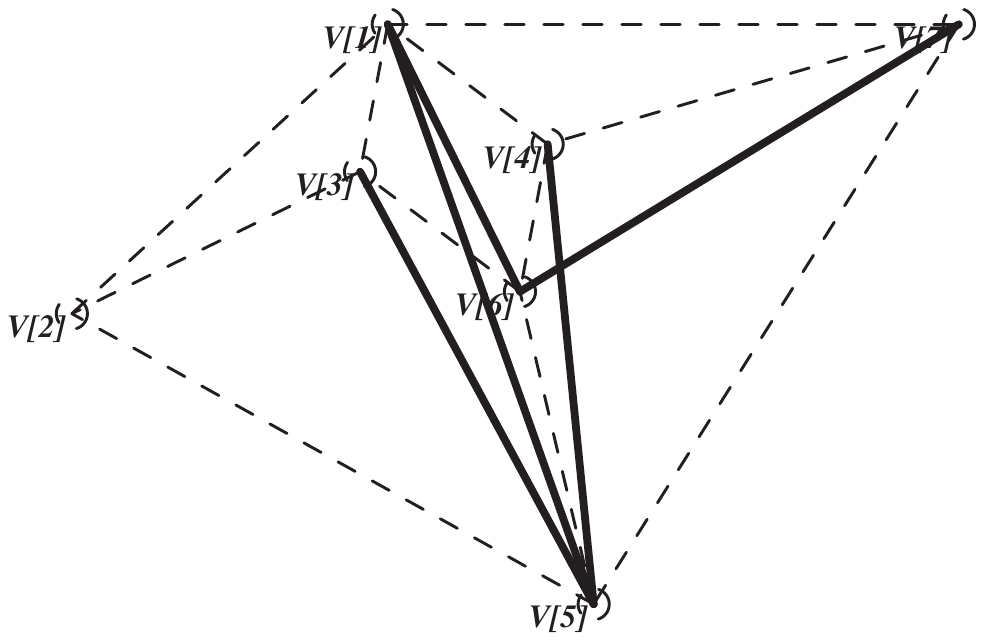}
    \vspace{-1cm}
\end{center}
\caption{When the edge lengths [15], [16], [35], [45] and [67] are specified as a solution of System (\ref{Esystem5x5}), the 11-bar linkage has only one assembly mode.} \label{11barLinkageDual}
\end{figure}

However, it is possible to find certain $5\times 5$ subsystems whose
subgraph is Laman and, moreover, uniquely define the configuration of the
overall linkage (Figure~\ref{11barLinkageDual}).
One of these systems has~4 bivariate equations and one trivariate
equation, and is defined by taking the following diagonal minors:
\begin{equation}\label{Esystem5x5}
    \left\{
    \begin{aligned}
        D(4, 5, 6, 7)(c_{46},c_{47},c_{56},c_{57},x_{45},x_{67}) = 0\\
        D(1, 4, 6, 7)(c_{14},c_{17},c_{46},c_{47},x_{16},x_{67}) = 0\\
        D(1, 4, 5, 7)(c_{14},c_{17},c_{47},c_{57},x_{15},x_{45}) = 0\\
        D(1, 2, 3, 5)(c_{12},c_{13},c_{25},c_{23},x_{15},x_{35}) = 0\\
        D(1, 3, 5, 6)(c_{13},c_{36},c_{56},x_{15},x_{16},x_{35}) = 0
    \end{aligned}
    \right.
\end{equation}
defines a system of~3 quadratic and two cubic equations in
$x_{15},x_{16},x_{35},x_{45},x_{67}$; the cubics are the first and last polynomials.
Now, this system's mixed volume turns out to be~56,
computed using the software from~\cite{EmCa95}
\footnote{http://www.di.uoa.gr/$\sim$emiris/}.
This bounds the number of (complex) common system's solutions, hence 
the number of (real) assembly modes.

Notice that this bound does not take into account solutions with zero
coordinates, in other words some zero length.
However, a linkage has assembly modes with some zero length only when the
input bar lengths form a singular set, in the sense that they would satisfy a
non-generic algebraic dependency. For example, by letting some input distance
be exactly $0$, some $11$-bar linkage may theoretically have infinitely many 
configurations. However, generically, it is impossible to have such a linkage.

Thus, the mixed volume bound guarantees that a true $11$-bar linkage
can have at most $56$ assembly modes.

\section{Lower bound}
\label{Slower}

To prove that 56 is a lower bound on the number of configurations,
it is sufficient to exhibit a linkage with 56 configurations.
Unfortunately, we have 11 design parameters. Thus our search space
is homeomorphic to $\RR_+^{11}$. Even if we reduce the search space
to the integers between 1 and 100 for each parameter, an exhaustive
search would lead us to consider $10^{22}$ linkages.

Another approach was used in \cite{LazMer94} to find a planar parallel
robot with 12 real solutions. The authors found a singular
positions with many real solutions, and use deformations to get the
desired result. The advantage of this approach is that the space of
singular configurations has a lower dimension than the full design
space. However, in our case, the dimension was still too high and
this approach could not allow us to conclude.

\subsection{Stochastic methods}

Random sampling is a common approach to search in high dimensional
space is random sampling. The system at hand being homogeneous, we
can specify the last coordinate $l_{10}$ to $100$ without
restriction of generality. Moreover, in the following, we restrict
our domain to integer coordinates.

Our goal is to maximize the function:
\[\begin{aligned}
&N :& {\mathbb{N}}^{10} & \rightarrow \mathbb{N} \notag\\
&   & (l_0,...,l_{9}) & \mapsto \begin{cases}
	0,\mbox{ if system~(\ref{Esystem5x5}) has infinitely many solutions.}
\\
	\mbox{Number of solutions of system~(\ref{Esystem5x5}), otherwise.}
                                       \end{cases}
\end{aligned}\]

A first idea is to sample random points using a Gaussian law
centered on an arbitrary initial point. This naive approach did not
allow us to find a linkage with 56 configurations, but we could
finally find such a point by using more sophisticated methods.

\subsubsection{Evaluation of the objective function}

The cost of evaluating the objective function is the bottleneck of
the stochastic methods. The function $N$ needs to be evaluated
several hundreds of time before converging toward a point where it
is maximal.

In our case, evaluating $N$ means solving a system of polynomial
equations with finitely many solutions, and counting the number
of its real positive roots. Experimentally, we use the Maple
function {\texttt{RootFinding[Isolate]}}. This function computes
isolating boxes around the real solutions of the system to solve.
The underlying algorithm uses Gröbner bases and Rational
Univariate Representation.

The choice of the system modeling the
11-bar manipulator is critical. A naive modelling yields a system
with $14$ variables, one for each of the coordinates of the $7$
vertices, and $11$ equations, one for each link in the
manipulator. In Maple, solving this system is roughly $10$
times slower than solving System (\ref{Esystem5x5}). That can be
explained partly by the fact that, generically, the time
complexity to solve a 0-dimensional system is roughly exponential
in the number of variables, e.g.\ \cite{CLO2,Lsigsam01}.

\subsubsection{Simulated annealing}

The Monte-Carlo simulation and its simulated annealing variant
\cite{KGVscience83} are the most well spread stochastic
optimisation methods. These methods have already been used in
robotics for path planning in \cite{KSLOieee96}. The convergence of
these methods has been well studied and the simulated annealing
simulations has been successfully used in different field such as
biology and chemistry. We implemented this method to maximize the
function $N$.

The Monte-Carlo simulation depends on a parameter $T$
called temperature. In the simulated annealing variant, the
temperature $T$ is decreased according to a specific schedule at
each step. The optimal way to decrease the temperature depends on
each problem. In our case, we chose arbitrarily a linear cooling
schedule (\cite{SHpla87,NAjpa98}). The corresponding pseudo-code is
summarized in Algorithm \ref{algorithm simulated annealing}.

\subsubsection{Cross entropy method}

A new method was introduced recently by Rubinstein in
\cite{Rejor97} for the simulation of rare events. This method is
especially well-suited for combinatorial and continuous
optimization (\cite{Rmcap99}). The idea of this approach consists
in minimizing the distance between specific probability laws
appearing during the computation (see \cite{BKMRaor05} for more
details). For our problem, we use the scheme developed for
continuous multi-extremal function optimization presented in
\cite{KPRmcap06}. The function $N$ is not continuous, but behaves
smoothly: generically, if we modify the lengths of a given
linkage by small enough values, the number of configurations is
increased or decreased only by $2$. For our problem it yields
Algorithm \ref{algorithm ce}.

\subsection{Results}

In Table~\ref{table:stochastic}, we compare the results of three
stochastic methods used to optimize the function $N$ with respect
to the lengths of the 11-bar linkage.

The first column shows the result of the direct random sampling.
Each line correspond to $600$ evaluation of $N$ on points chosen
according to a Gaussian law of center $100$ and standard deviation
$100$. This operation has been run $10$ times, yielding a total of
$6000$ evaluations of $N$ on random points. However, this did not
allow us to find a linkage with $56$ configurations.

The second approach is a Simulated Annealing simulation. We ran it
$10$ times, limiting the number of evaluations of $N$ to $600$. One
of our simulation returned a set of link lengths yielding a
linkage with $56$.

Finally, the last column reports the results of the Cross Entropy
Method. Each line correspond to a simulation stopped after $600$
evaluations of $N$. This approach was the most efficient. Four
simulations out of ten returned a linkage with $56$
configurations.

\begin{figure}[htb] \centering
\includegraphics[angle=90,width=10.5cm]{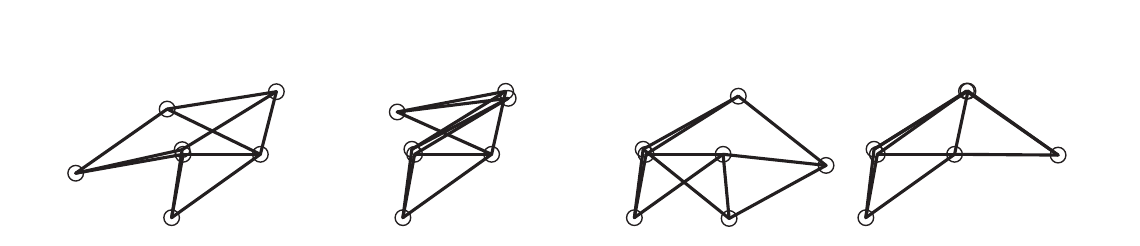}\\
\includegraphics[angle=90,width=10cm]{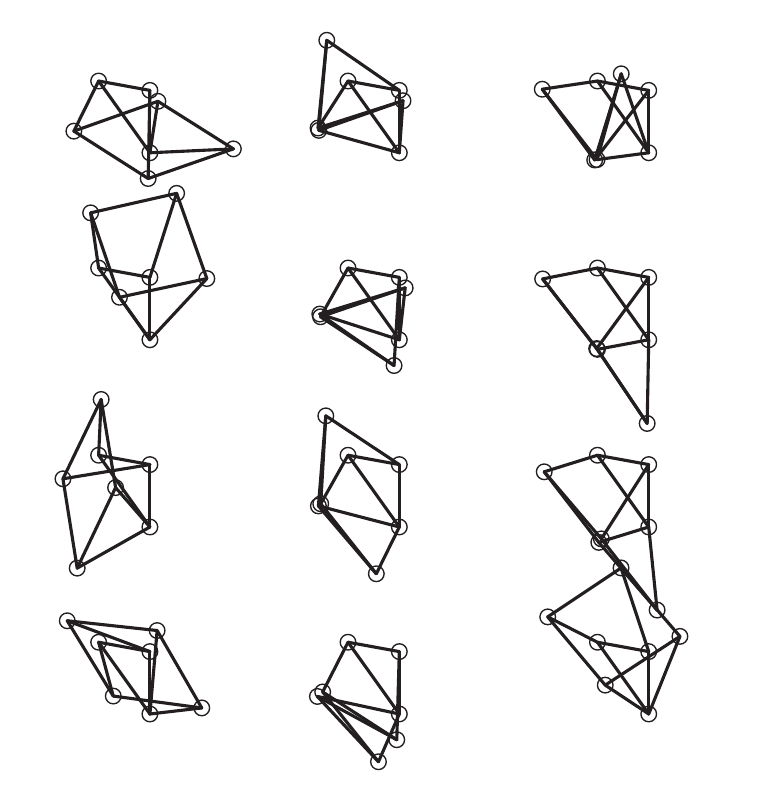}\\
\includegraphics[angle=90,width=10cm]{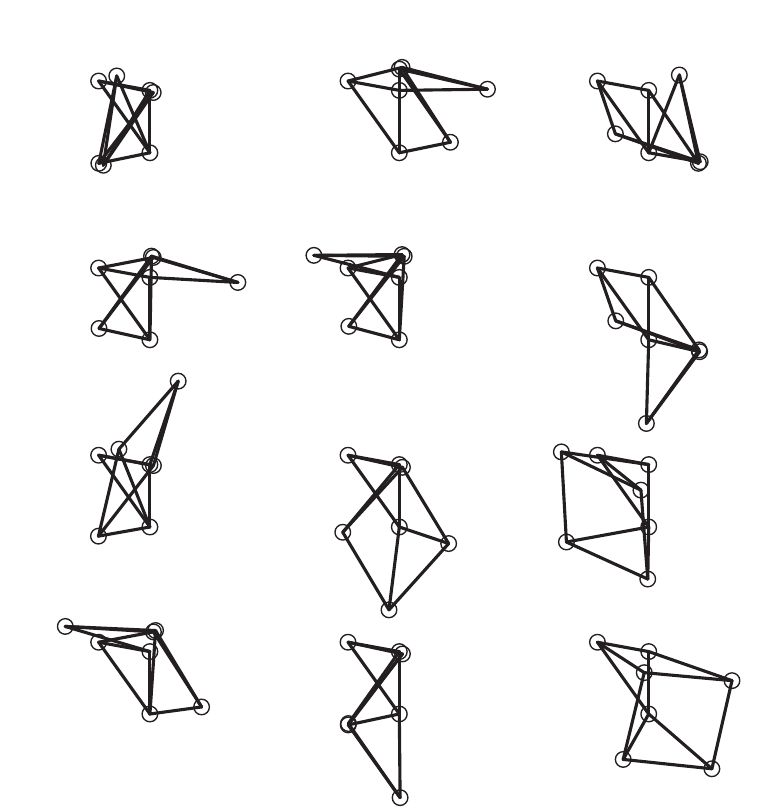}
\caption{28 of the 56 configurations of our linkage.
The other 28 can be deduced by symmetry with respect to the horizontal axis.\label{F_26configs}}
\end{figure}

The results of our stochastic methods are summarized in Table
\ref{table:stochastic}. In particular, we found that the
manipulator with the following design parameters

\[\begin{aligned}
   l_0 &= 180 & l_1 &= 70      & l_2 &= 200 & l_3 &= 205 \\
   l_4 &= 210 & l_5 &= 205     & l_6 &= 80  & l_7 &= 200 \\
   l_8 &= 70  & l_9 &= 200     & l_{10} &= 100
\end{aligned}\]

has exactly 56 assembly modes. Some of its assembly modes are shown
in Figure \ref{F_26configs}.

Our simulations show that the set of $11$-bar linkages with $56$ assembly modes
has a non-zero volume. With the direct random sampling,
we did not find any linkage with more than $48$ assembly modes. This indicates that linkages with bar lengths between $0$ and $300$ have a very small probability to have $56$ assembly modes. Our experiments show that the Cross Entropy method is a general approach well suited to generate such configurations. Moreover, it can be used easily for other mechanisms, such as larger planar or spatial linkages for example. A variant of this method can also be used to compute an estimation of the probability to find a linkage with a maximal number of assembly mode.

\begin{table}[h]
\begin{center} \begin{tabular}{ccc}
Direct Sampling & Simulated Anealing & Cross Entropy
\\
\hline
44 (572)
 & 52 (17)
 & 52 (199)
\\
42 (196)
 & 54 (247)
 & 54 (132)
\\
48 (27)
 & 48 (362)
 & 52 (186)
\\
44 (200)
 & 52 (14)
 & 54 (130)
\\
42 (200)
 & 54 (547)
 & 56 (497)
\\
44 (424)
 & 54 (315)
 & \textbf{56} (328)
\\
46 (48)
 & \textbf{56} (425)
 & \textbf{56} (454)
\\
42 (170)
 & 50 (585)
 & 54 (375)
\\
42 (18)
 & 54 (26)
 & \textbf{56} (552)
\\
46 (366)
 & 52 (474)
 & \textbf{56} (355)
\\
\end{tabular}
\end{center}
\caption{Results of different stochastic optimisation algorithms. Each line corresponds to a simulation stopped after 600 evaluations of $N$. The results is of the form $n (m)$ where $n$ is the maximal reached value of $N$ and $m$ is the number of evaluations done before reaching it. \label{table:stochastic}}
\end{table}

\begin{algorithm}
\caption{Simulated annealing with linear cooling to maximize the
number of solutions of System (\ref{Esystem5x5})}
\label{algorithm simulated annealing}
\begin{algorithmic}
\STATE \COMMENT{Simulated annealing parameters.}
\STATE $T_0 \gets 4$
\STATE $maxStep \gets 1000$
\STATE $\vec{\sigma} \gets [10,...,10]$

\COMMENT{Initial point and number of configuration.}
\STATE $point \gets \{l_0=100, ..., l_{9}=100\}$
\STATE $value \gets N (l_0,...,l_{9})$

\FOR {$n$ from 1 to $maxStep$ while $value<56$}
    \STATE\COMMENT{Compute a neighbour point with normal laws of variances
                   $\sigma_i^2$ and centered on the coordinates of the
                   previous point.}
    \STATE $newPoint \gets GaussianNeighbour (point, \vec{\sigma})$

    \COMMENT{Compute the number of configuration.}
    \STATE $newValue \gets N (l_0,...,l_{9})$

    \STATE $treshold \gets$ Uniform random value in [0,1]

    \COMMENT{Acceptance of the new value.}
    \IF {$\left\{\begin{aligned} newValue &> value\\
                        &or\\
                 treshold &< e^{\frac{newValue-value}{T}}
          \end{aligned}\right.$}
        \STATE $point \gets newPoint$
        \STATE $value \gets newValue$
    \ENDIF
    \STATE $T \gets T_0(1 - n/maxStep)$
\ENDFOR

\RETURN point
\end{algorithmic}

\end{algorithm}

\begin{algorithm}
\caption{Cross Entropy Method to maximize the number of solutions of
         System (\ref{Esystem5x5})}
\label{algorithm ce}
\begin{algorithmic}
    \STATE\COMMENT{Cross Method parameters.}
    \STATE $maxStep \gets 600$
    \STATE $N_{elite} \gets 5$
    \STATE $N \gets 20$
    \STATE $\alpha \gets 0.5$

    \COMMENT{Initial point and number of configuration.}
    \STATE $\vec{\mu} \gets [100,...,100]$
    \STATE $\vec{\sigma} \gets [100,...,100]$
    \STATE $\gamma \gets 0$
    \STATE $max \gets 0$

    \FOR {$n$ from $1$ to $maxStep$ while $max<56$}
        \FOR {$i$ from $1$ to $N$}
            \STATE $\mathbf{X_i} \gets GaussianNeighbour(\vec{\mu},\vec{sigma})$
        \ENDFOR
        \STATE $prevmax \gets max$

        \COMMENT{Evaluate the function $N$ on each sample point.}
        \FOR {$i$ from $1$ to $N$}
            \STATE $\mathbf{V}_i \gets N(\mathbf{X_i})$
            \IF {$\mathbf{V_i}>max$}
                \STATE $max \gets \mathbf{V}_i$
                \STATE $point \gets \mathbf{X_i}$
            \ENDIF
        \ENDFOR;
        \STATE Sort $\mathbf{X}$ and $\mathbf{V}$ from the largest
               $\mathbf{V}_i$ to the smallest

        \COMMENT{Compute the new Gaussian laws.}
        \STATE $\gamma \gets \mathbf{V}_{N_{elite}}$
        \STATE $\vec{\mu}^{new} \gets \frac{1}{N_{elite}}\sum_1^{N_{elite}}
                                  \mathbf{X_i}$
        \STATE $\vec{\sigma}^{new} \gets \sqrt{\frac{1}{N_{elite}}\sum_1^{N_{elite}}
                                  (\mathbf{X_i}-\vec{\mu}_i)^2}$

        \COMMENT{Smooth updating.}
        \STATE $\vec{\mu} \gets \alpha\vec{\mu}^{new} + (1-\alpha)\vec{\mu}$
        \STATE $\vec{\sigma} \gets \alpha\vec{\sigma}^{new} +
                                   (1-\alpha)\vec{\sigma}$
    \ENDFOR
    \RETURN point
\end{algorithmic}
\end{algorithm}

\section{Improvement on the general lower bound}

This specific linkage allows us to improve slightly the lower
bound on the maximal number of assembly modes of a planar linkage
with $n$ pivot joints. We use the same fan construction as in
\cite{BS04}.

We construct a linkage with $n$ nodes composed of $k$
sub-mechanisms as follows. Each of the $k$ sub-mechanisms is a
11-bar linkage with the topology we have considered so it has 56
assembly modes, and 28 when one triangle is fixed.
All sub-mechanisms share this triangle so there are
no degrees of freedom other than those within each sub-mechanism,
and the overall mechanism is rigid. The total number of
assembly modes equals the product of remaining assembly modes per
sub-mechanism, since these sub-mechanisms can be configured
independently. Hence the overall number is $28^k$ when one triangle
is specified, and $2 \cdot 28^k$ in total. For $n$ pivot joints, we
let $k$ be the greatest integer $\le\frac{n-3}{4}$. This
yields the lower bound:
\[2\cdot 28^{\lfloor \frac{n-3}{4} \rfloor}\]

The constant under the exponent is $\sqrt[4]{28}\simeq 2.3003$
while in the previous lower bound the constant was
$\sqrt[3]{12}~\simeq~2.2894$.

\section{Further work}

Undoubtedly, the most important and oldest problem in rigidity theory
is the full combinatorial characterization of rigid graphs in $\RR^3$.
We believe that one can extend our methods to spatial linkages.
We expect our approach leads to an algorithmic process for obtaining good
algebraic representations of the enumeration problem at hand, namely
low mixed volumes for the systems derived from distance matrices.
One issue is that the number
of equations produced is quite large, with algebraic dependancies among them. 
The question becomes then to choose the best well-constrained system.

The structures studied in this paper are point-and-bar structures; they
generalize to body-and-bar, where edges can be connected to different
points of a rigid body.
It is known that a body-and-bar structure in $\RR^d$ is rigid if and only if
the associated graph 
is the edge-disjoint union of ${d+1\choose 2}$ spanning trees~\cite{WT85}.
Body-and-bar structures are our next object of study.

\bigskip

\noindent
\textbf{Acknowledgements.}
I.Z.~Emiris is partially supported by FP7 contract PITN-GA-2008-214584
SAGA: Shapes, Algebra, and Geometry.
He also acknowledges fruitful discussions with Elias Tsigaridas.
Both authors were inspired to work on this problem at the Workshop on
Discrete and Algebraic Geometry in September 2010, at Val d'Ajol, France.


\psfrag{V[1]}[Br][tr]{\raisebox{-0.5em}{$1$}}
\psfrag{V[2]}[bl][cr]{\raisebox{0.2em}{$\ 2$}}
\psfrag{V[3]}[bl][cc]{\raisebox{0.4em}{$\qquad\ \ 3$}}
\psfrag{V[4]}[Br][tr]{$4$} \psfrag{V[5]}[Br][tr]{$5$}
\psfrag{V[6]}[Bl][tr]{\raisebox{-0.7em}{$\ \ \  6$}}
\psfrag{V[7]}[Bl][tr]{\raisebox{0.7em}{$7$}}

\end{document}